\documentclass{llncs}
\usepackage{makeidx}  

\usepackage{epsfig}
\usepackage{algorithm}
\usepackage{algorithmic}
\usepackage{amsmath}
\usepackage{amssymb}
\usepackage{tabularx}
\usepackage{hyperref}
\DeclareMathSymbol{\arra}{\mathbin}{AMSb}{'122}

\begin{document}
\frontmatter          
\pagestyle{headings}  
\addtocmark{Fixed-sized clusters $k$-Means} 
%

%
\mainmatter              
\title{Fixed-sized clusters $k$-Means}
\titlerunning{Fixed-sized clusters $k$-Means}  
%
\author{Mikko I. Malinen \and Pasi Fr\"anti
}
\authorrunning{Mikko I. Malinen and Pasi Fr\"anti} 
%
\tocauthor{Mikko I. Malinen, Pasi Fr\"anti}
\institute{School of Computing, University of Eastern Finland, Box 111, FIN-80101 Joensuu, Finland,\\
\email{mmali@cs.uef.fi, franti@cs.uef.fi},\\ WWW home page:
\texttt{http://cs.uef.fi/\homedir mmali, http://cs.uef.fi/pages/franti}
}

\maketitle              

\begin{abstract}
We present a $k$-means-based clustering algorithm, which optimizes the mean square error, for given cluster sizes. A straightforward application is balanced clustering, where the sizes of each cluster are equal. In the $k$-means assignment phase, the algorithm solves an assignment problem using the Hungarian algorithm. This makes the assignment phase time complexity $O(n^3)$. This enables clustering of datasets of size more than 5000 points.
\end{abstract}
\section{Introduction} \label{se:introduction}
Euclidean sum-of-squares clustering is an NP-hard problem~\cite{euclclustNPhard}, which groups $n$ data points into $k$ clusters so that intra-cluster distances are low and inter-cluster distances are high. Each group is represented by a center point (centroid). The most common criterion to optimize is the mean square error (MSE):
\begin{equation}
\text{MSE}= \sum _{j=1}^k \sum _{X_i\in C_j}\frac{\mid \mid X_i-C_j \mid \mid ^2}{n}, 
\end{equation} 
where $X_i$ denotes data point locations and $C_j$ denotes centroid locations. $k$-Means~\cite{kmeansmacqueen} is the most commonly used clustering algorithm, which provides a local minimum of MSE given the number of clusters as input. $k$-Means algorithm consists of two repeatedly executed steps:\\
{\bf Assignment step:} Assign the data points to clusters specified by the nearest centroid:
$$P_j^{(t)}= \{ X_i : \| X_i - {C}_j^{(t)}\| \le \| X_i -{C}_{j^*}^{(t)}\|$$
$$\forall \quad  j^* = 1,...,k\}$$ 
{\bf Update step:} Calculate the mean of each cluster:
$${C}_j^{(t+1)}= \frac{1}{|P_j^{(t)}|}\sum _{X_i \in P_j^{(t)}}X_i$$
These steps are repeated until the centroid locations do not change anymore. $k$-Means assignment step and the update step are optimal with respect to MSE: The partitioning step minimizes MSE for a given set of centroids; the update step minimizes MSE for a given partitioning. The solution therefore converges to a local optimum but without guarantee of global optimality. To get better results than in the $k$-means, slower agglomerative algorithms~\cite{PNN,FastAgglomkNN,IterativeShrinking} or more complex $k$-means variants~\cite{kmeansplusplus,randomswap,xmeans,globalkmeans} are sometimes used.

In \emph{balanced clustering}~\cite{balancedkmeansssspr14,All-pairwisesquareddistances}, there are equal or more equal number of points in each cluster than in traditional clustering. Balanced clustering is desirable, for example, in divide-and-conquer methods, where the divide step is done by clustering. 

Balanced clustering, in general, is a 2-objective optimization problem in which two goals contradict each other: to minimize MSE and to balance cluster sizes. Traditional clustering aims to minimize MSE without considering cluster size balance. Balancing, on the other hand, would be trivial if we did not care about MSE; simply by dividing points to equal size clusters randomly.\\ 
\ \\
We next review some articles that have size constraits on clusters.
Constrained $k$-means~\cite{Constrainedkmeans} allows putting lower bound on cluster sizes. Data clustering with size constraints~\cite{sizeconstraints} transforms the problem into a binary integer linear programming problem. The biggest dataset in their experiments is of size 625 points, indicating that the algorithm is not suitable for bigger datasets. Data Clustering with Cluster Size Constraits~\cite{DataClusteringwithClusterSize} allows putting upper bounds on cluster sizes. Their biggest dataset tested is 2000 points that also indicates that bigger datasets take too much time. Our proposed algorithm allows clustering up to circa 5000 points.\\

\section{Fixed-sized clusters $k$-means}
\begin{figure*}
\begin{center}
\epsfig{file=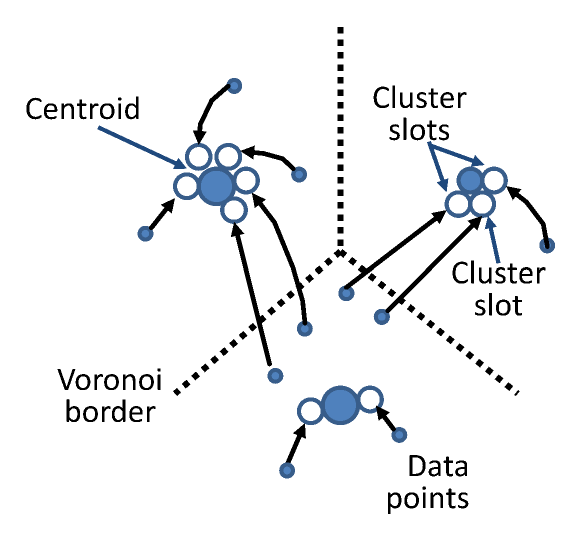,width=7cm} 
\caption{Assigning points to centroids via cluster slots.} \label{fi:clusterslots}
\end{center}
\end{figure*}

To describe Fixed-sized clusters $k$-means, we need to define what is an assignment problem. The formal definition of assignment problem (or linear assignment problem) is as follows. Given two sets ($A$ and $S$), of equal size and with a weight function $W:A\times S \rightarrow \arra$. The goal is to find a bijection $f:A\rightarrow S$ so that the cost function is minimized:
$$\text{Cost} = \sum _{a\in A}W(a,f(a)).$$
In the context of the proposed algorithm, sets $A$ and $S$ correspond respectively to cluster slots and to data points, see Figure~\ref{fi:clusterslots}.

In Fixed-sized clusters $k$-means, we proceed as in $k$-means, but the assignment phase is different: Instead of selecting the nearest centroids we have $n$ pre-allocated slots, and datapoints can be assigned only to these slots; see Figure~\ref{fi:clusterslots}. 
\begin{figure*}
\begin{center}
\epsfig{file=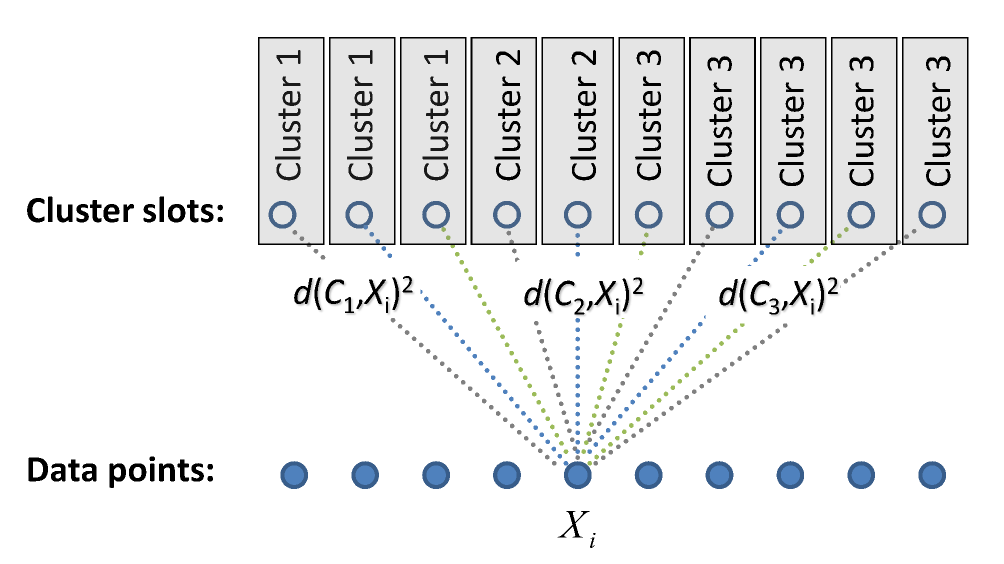, width=10cm}
\caption{Minimum MSE calculation with fixed-sized clusters. Modeling with bipartite graph.} \label{fi:assignmentgraph}
\end{center}
\end{figure*}

To find an assignment that minimizes MSE, we solve an assignment problem using the Hungarian algorithm~\cite{AssignmentProblBook}. First we construct a bipartite graph consisting $n$ datapoints and $n$ cluster slots, see Figure~\ref{fi:assignmentgraph}. We then partition the cluster slots in clusters in fixed sizes.

We give centroid locations to partitioned cluster slots, one centroid to each cluster. The initial centroid locations can be drawn randomly from all data points. The edge weight is the squared distance from the point to the cluster centroid it is assigned to. Contrary to standard assignment problem with fixed weights, here the weights dynamically change after each $k$-Means iteration according to the newly calculated centroids. After this, we perform the Hungarian algorithm to get the minimal weight pairing. The squared distances are stored in a $n\times n$ matrix, for the sake of the Hungarian algorithm. The update step is similar to that of $k$-means, where the new centroids are calculated as the means of the data points assigned to each cluster: 
\begin{equation} \label{eq:updatestep}
C_i^{(t+1)} = \frac{1}{n_i} \cdot \sum _{X_j\in C_i^{(t)}} X_j.
\end{equation} 
The weights of the edges are updated immediately after the update step. The pseudocode of the algorithm is in Algorithm~\ref{al:balanced}. In calculation of edge weights, the cumulative sum of cluster sizes is 
\begin{equation}
    c(j)=\sum _{l=[1..j]}n_l\quad \forall j\in[1..k],
\end{equation}
where $n_l$:s are cluster sizes and the number of cluster slot is denoted by $a$ and
\begin{equation}
    {\arg\min}_j\ \ c(j)\ge a
\end{equation}
is used in calculation of cluster where a cluster slot belongs to. So the edge weights are calculated by
\begin{equation} \label{eq:calculationofweights}
W(a,i) = dist(X_i,C^t_{{\arg\min}_j\  c(j)\ge a})^2 \quad \forall a\in [1..n] \quad \forall i\in [1..n].
\end{equation}
After convergence of the algorithm the partition of points $X_i$, $i\in [1..n]$, is
\begin{equation}
X_{f(a)} \in P_{{\arg\min}_j\  c(j)\ge a}. 
\end{equation}
\begin{algorithm}
\caption{Fixed-sised clusters $k$-Means\newline Input: \quad\quad dataset $X$, cluster sizes $n_l$, number of clusters $k$\newline Output: \quad partitioning of dataset.} \label{al:balanced}
\begin{algorithmic}
\STATE Initialize centroid locations $C^0$.\\
\STATE $t\gets 0$
\REPEAT
\STATE Assignment step:\\
\STATE \qquad \qquad Calculate edge weights. Eq.~\ref{eq:calculationofweights}
\STATE \qquad \qquad Solve an Assignment problem.\\
\STATE Update step:\\
\STATE \qquad \qquad Calculate new centroid locations $C^{t+1}$. Eq.~\ref{eq:updatestep}\\
\STATE $t\gets t+1$
\UNTIL centroid locations do not change.\\
\STATE Output partitioning.
\end{algorithmic}
\end{algorithm}
\newpage
There is a convergence result in~\cite{Constrainedkmeans} (Proposition 2.3) for Constrained $k$-means. The result says that the algorithm terminates in a finite number of iterations at a partitioning that is locally optimal. At each iteration, the cluster assignment step cannot increase the objective function of Constrained $k$-means (3) in~\cite{Constrainedkmeans}. The cluster update step will either strictly decrease the value of the objective function or the algorithm will terminate. Since there are a finite number of ways to assign $m$ points to $k$ clusters so that cluster $h$ has at least $\tau _h$ points, since Constrained $k$-means algorithm does not permit repeated assignments, and since the objective of Constrained $k$-means (3) in~\cite{Constrainedkmeans} is strictly nonincreasing and bounded below by zero, the algorithm must terminate at some cluster assignment that is locally optimal. The same convergence result applies to Fixed-sized clusters $k$-means as well.   
The assignment step is optimal with respect to MSE because of pairing and the update step is optimal, because MSE is clusterwise minimized as is in $k$-means.

\section{Time Complexity}
Time complexity of the assignment step in $k$-means is $O(k\cdot n)$.  The assignment step of the proposed Fixed-sized clusters $k$-means algorithm can be solved in $O(n^3)$ time with the Hungarian algorithm. 

\section{Application: Seating plan}

As an application we present calculating a seating plan, where compatibility of persons within tables is optimized. First we need a compatibility matrix, where compatibility distance is given for every pair of persons. This has to be done manually.
\begin{equation}
D =
\begin{pmatrix}
0 & d_{12} & .&.&.\\
d_{21} & 0 & & &\\
. &   &. & & \\
. &   & & .&\\
. &   & & &.\\
\end{pmatrix}
\end{equation}
Then we need to do multidimensional scaling~\cite{coxandcox} giving $D$ as argument and the result is the data $X$ in higher dimensional space, but distances preserved.
Then we do Fixed-sized k-Means giving data $X$, sizes of tables $n_l$ and number of tables $k$ as arguments. The output is the seating plan.

\subsection{Experiments}

We tested the algorithm by creating a seating plan for Mikko I. Malinen's doctoral dissertation evening party in 2015. There were 22 persons invited. In compatibility distance matrix there were $22\cdot 22=484$ distances. Sizes of tables were $[4\ 4\ 5\ 6\ 3]$ and $k$ was 5. Data $X$ became 10-dimensional. We repeated the algorithm 1000 times. This took only few seconds. People were happy with the seating plan. The software for both Fixed-sized clusters $k$-means and Seating plan are available from \url{http://cs.uef.fi/~mmali/software/}.
\section{Conclusions}
We presented an algorithm for clustering giving cluster sizes as constraints. The algorithm is practical up to 5000 points data. As an application we presented creating a seating plan for f.eg. parties.
%

\bibliographystyle{splncs03}

\bibliography{Fixed_kmeans.bib}

%
%



\end{document}